\documentclass[conference]{IEEEtran}
\IEEEoverridecommandlockouts
\usepackage{cite}
\usepackage{amsmath,amssymb,amsfonts}
\usepackage{algorithmic}
\usepackage{graphicx}
\usepackage{textcomp}
\usepackage{xcolor}
\usepackage{subcaption}
\usepackage{comment}

\def\BibTeX{{\rm B\kern-.05em{\sc i\kern-.025em b}\kern-.08em
    T\kern-.1667em\lower.7ex\hbox{E}\kern-.125emX}}
\begin{document}

\title{HAL-NeRF: High Accuracy Localization Leveraging Neural Radiance Fields\\

{\footnotesize \textsuperscript{}}

\thanks{© 2025 IEEE. Personal use of this material is permitted. Permission from IEEE must be obtained for all other uses, in any current or future media, including reprinting/republishing this material for advertising or promotional purposes, creating new collective works, for resale or redistribution to servers or lists, or reuse of any copyrighted component of this work in other works. This preprint has not undergone any post submission improvements or corrections. This work was supported by the European Union’s Horizon Europe programme under grant number 101092875 “DIDYMOS-XR” (https://www.didymos-xr.eu). The computational resources were granted with the support of GRNET.}
}

\author{
Asterios Reppas, Grigorios-Aris Cheimariotis, Panos K. Papadopoulos, Panagiotis Frasiolas, Dimitrios Zarpalas
}

\maketitle

\begin{abstract}
Precise camera localization is a critical task in XR applications and robotics. Using only the camera captures as input to a system is an inexpensive option that enables localization in large indoor and outdoor environments, but it presents challenges in achieving high accuracy. Specifically, camera relocalization methods, such as Absolute Pose Regression (APR), can localize cameras with a median translation error of more than $\mathbf{0.5m}$ in outdoor scenes. This paper presents HAL-NeRF, a high-accuracy localization method that combines a CNN pose regressor with a refinement module based on a Monte Carlo particle filter. The Nerfacto model, an implementation of Neural Radiance Fields (NeRFs), is used to augment the data for training the pose regressor and to measure photometric loss in the particle filter refinement module. HAL-NeRF leverages Nerfacto's ability to synthesize high-quality novel views, significantly improving the performance of the localization pipeline. HAL-NeRF achieves state-of-the-art results that are conventionally measured as the average of the median per scene errors. The translation error was $\mathbf{0.025m}$ and the rotation error was $\mathbf{0.59^{\circ}}$ and $\mathbf{0.04m}$ and  $\mathbf{0.58^{\circ}}$ on the 7-Scenes dataset and Cambridge Landmarks datasets respectively, with the trade-off of increased computational time. This work highlights the potential of combining APR with NeRF-based refinement techniques to advance monocular camera relocalization accuracy.

\end{abstract}

\begin{IEEEkeywords}
Camera relocalization, NeRF, Pose Regression, Pose Refinement
\end{IEEEkeywords}

\section{Introduction}
\label{intro}

Accurate localization of a robot or camera is crucial for extended reality applications, robotics \cite{lim2015realtime} and autonomous driving \cite{wen2022tm3loc}, as it enables precise interaction with the environment. An option of choice is the estimation of a camera's position and orientation by its corresponding 2D capture within a specified coordinate system, typically referenced to a pre-computed map. This task, commonly named monocular camera relocalization, does not include the use of more expensive sensors (lidar, rgb-d).  

Various computational methods have been developed for directly inferring poses from images. A prominent group of techniques is Absolute Pose Regression (APR) \cite{kendall2016modelling}, \cite{kendall2017geometric}, \cite{shavit2021paying}, \cite{kendall2015posenet}, which directly predicts camera poses from input images using deep learning models. While APR methods offer the advantage of direct pose estimation with lower computational resource requirements, they often face challenges in achieving high accuracy due to factors such as limited training data and complex scene variations.

Other methods leverage end-to-end architectures to extract pose-related information from implicit representations \cite{krizhevsky2012imagenet}, \cite{ren2016faster}, \cite{tang2023neumap}. A critical aspect of all image-based localization methods is the construction of a robust map that can effectively support the localization pipeline. The quality of the map is essential for ensuring accurate and reliable position estimation. 

Recently, the rapid advancements in Neural Radiance Fields (NeRFs), a modern implicit scene representation, have positioned them as a widely adopted component in localization pipelines. NeRFs enable the synthesis of novel views from the learned scene representations. This capability is useful for augmenting training datasets. In \cite{moreau2022lens}, Nerf in the Wild \cite{martin2021nerf} was used to augment a pose regressor. Besides, Loc-NeRF \cite{maggio2023loc} used photometric loss between NeRF-rendered views and a query image to update a particle filter optimization scheme.

This paper proposes a high-accuracy localization pipeline, which has two main modules. The first module is a CNN-pose regressor, used also by \cite{chen2022dfnet}, to extract an initial pose estimation. Its training used a dataset that is augmented with rendered views from Nerfacto \cite{tancik2023nerfstudio}. The second module is a test-time refinement step that used a Monte Carlo particle filter inspired by \cite{maggio2023loc}. The main modification of the proposed method, compared to \cite{chen2022dfnet}, was replacing the training of Vanilla NeRF \cite{mildenhall2021nerf} with Nerfacto \cite{tancik2023nerfstudio}. Additionally, the particle filter was employed as a refinement method based on an initial prediction, rather than being used for global localization. NeRFs were used in both cases to calculate the particle weight updates in an iterative refinement process.

Adopting a more advanced NeRF model (Nerfacto) \cite{tancik2023nerfstudio} in the proposed localization pipeline significantly improved accuracy. The improved quality of the synthesized views, and the computational efficiency of rendering, enhanced the overall performance and adaptability of the localization pipeline to various scenes. 

The main contribution of this paper is the merging of CNN-pose regressor with a particle filter pose refinement module. The regressor and the refinement module leverage a high-quality NeRF. The whole pipeline achieved more accurate pose predictions in most scenes of the Cambridge Landmarks and 7-scenes datasets. 

This paper is structured as follows: In Section~\ref{relatedwork}, related work on camera relocalization and the integration of NeRFs in localization pipelines is reviewed. Section~\ref{method} describes the HAL-NeRF pipeline. Section~\ref{implementation} outlines the implementation details. Experimental results on the 7-Scenes and Cambridge Landmarks datasets are presented in Section~\ref{results}, demonstrating the effectiveness of the proposed method. Finally, Section~\ref{discussion} discusses the advantages and limitations of our approach, followed by conclusions and future directions in Section~\ref{conclusion}.

\section{Related Work}
\label{relatedwork}

\subsection{Absolute Pose Regression}

 An APR takes an image as input, processes it through convolutional layers, and outputs the camera's position and orientation. The key advantage of APRs is their ability to deliver fast, direct pose estimates from visual data, making them highly efficient for real-time applications. The pioneering model in this category was introduced in \cite{kendall2015posenet}. PoseNet employs GoogLeNet, a 22-layer convolutional network, as its backbone to regress the 6-DOF camera pose from a single RGB image in an end-to-end fashion. Since its introduction, PoseNet has undergone several architectural modifications. These include the addition of LSTM layers for improved sequential processing \cite{walch2017image}, the incorporation of Monte Carlo dropout into a Bayesian CNN to estimate pose with uncertainty \cite{kendall2016modelling}, and the adoption of geometrically-oriented loss functions, such as reprojection loss \cite{kendall2017geometric}, to enhance pose accuracy. MapNet \cite{brahmbhatt2018geometry} extends the approach by learning a deep neural network by simultaneously minimizing the loss for both the absolute pose of individual images and the relative pose between image pairs. In \cite{shavit2021learning}, they introduce an approach for multi-scene absolute camera pose regression using Transformers. This method employs encoders to aggregate activation maps through self-attention mechanisms, while decoders convert latent features and scene encodings into potential pose predictions. Finally, Camera Pose Auto-Encoders (PAEs) \cite{shavit2022camera} are introduced as a method to improve camera pose estimation by encoding pose representations using a Teacher-Student approach, with Absolute Pose Regressors (APRs) serving as the teachers.

 APR methods are capable of localizing cameras but with relatively low accuracy in challenging outdoor scenes. Specifically, (PAEs) \cite{shavit2022camera} achieve $0.96m$ median translation error and $2.73^{\circ}$ median rotational error in the Cambridge Landmakrs test case. Although, there are limitations in the 3D scenes, related to incomplete capture and transient objects, NeRF-based methods demonstrate that there is room for improvement in localization.
 
\subsection{NeRFs in Localization}

The numerous advantages of NeRFs have established them as a critical component in localization algorithms. Beginning with an initial pose estimate, iNeRF \cite{yen2021inerf} can use a NeRF model to estimate 6 DoF pose for scenes and objects with complex geometry by applying gradient descent to minimize the discrepancy between pixels rendered by a NeRF and those in the observed image. LATITUDE \cite{zhu2023latitude} follows a similar approach to iNeRF by incorporating an initial pose prediction; however, it diverges in the optimization stage by employing a Truncated Dynamic Low-pass Filter, which helps prevent the optimization process from becoming trapped in local optima. Leveraging Monte Carlo localization as a core mechanism for pose estimation with a NeRF map model, Loc-NeRF \cite{maggio2023loc} achieves faster localization and operates without the need for an accurate initial pose estimate. LENS \cite{moreau2022lens} applies novel view synthesis to the robot relocalization problem, demonstrating enhanced camera pose regression accuracy by incorporating an additional synthetic dataset rendered using NeRFs. DFNet \cite{chen2022dfnet} also applies data augmentation using NeRFs and further distinguishes itself by leveraging direct feature matching, in contrast to the conventional photometric loss approaches. This additional strategy enhances pose regression accuracy through semi-supervised learning on unlabeled data.

The aforementioned methods were among the first to be introduced in this domain. Since their inception, researchers aim to fully exploit the pose information provided by NeRF's implicit representation.
NeFeS \cite{chen2024neural} proposes a test-time refinement pipeline that enhances APR methods by leveraging implicit geometric constraints through a robust feature field, enabling better utilization of 3D information during inference. Additionally, it introduces the Neural Feature Synthesizer (NeFeS) model, which encodes 3D geometric features during training and directly renders dense novel view features at test time to refine APR methods. A similar approach is presented in \cite{lin2024learning}, where a novel neural volumetric pose feature, termed PoseMap, is introduced. However, in that work, PoseMap is integrated directly into the APR model rather than being used as a refinement method. CROSSFIRE \cite{moreau2023crossfire} presented an implicit representation for local descriptors, facilitating iterative 2D-3D feature matching. Finally, NerfMatch \cite{zhou2024nerfect} follows a similar approach to CROSSFIRE.

\subsection{NeRF - Overview}

NeRF represents a scene as a continuous volumetric function parameterized by a Multi-Layer Perceptron (MLP). Specifically, the 3D position $\mathbf{r}(t)$ and viewing direction $\mathbf{d}$ along a camera ray $\mathbf{r}(t) = \mathbf{o} + t\mathbf{d}$, are passed as inputs to an MLP with weights $\Theta$ to compute, for each sample along the ray, the volume density $\sigma$ and the RGB color $\mathbf{c}$ corresponding to the radiance emitted.

\begin{equation}
    \sigma(t), \mathbf{c}(t) = \text{MLP}_{\Theta}(\mathbf{r}(t), \mathbf{d}).
\end{equation}

A key design decision made in NeRF is to structure the MLP such that volume density is only predicted as a function of 3D position, while emitted radiance is predicted as a function of both 3D position and 2D viewing direction.

To render the color $\hat{\mathbf{C}}(\mathbf{r})$ of a pixel, NeRF queries the MLP at sampled positions $t_k$ along the corresponding ray and uses the estimated volume densities and colors to approximate a volume rendering integral using numerical quadrature, as discussed by \cite{max1995optical}:

\begin{equation}
    \hat{\mathbf{C}}(\mathbf{r}) = \sum_{k} T(t_k) \alpha(\sigma(t_k)\delta_k) \mathbf{c}(t_k),
\end{equation}

where

\begin{equation}
    T(t_k) = \exp\left( -\sum_{k'=1}^{k-1} \sigma(t_{k'}) \delta_{k'} \right), \quad \alpha(x) = 1 - \exp(-x),
\end{equation}

and $\delta_k = t_{k+1} - t_k$ is the distance between two adjacent points along the ray.

NeRF trains the MLP by minimizing the squared error between input pixels from a set of observed images (with known camera poses) and the pixel values predicted by rendering the scene as described above:

\begin{equation}
    \mathcal{L}_r = \sum_{i} \left\lVert \mathbf{C}(\mathbf{r}_i) - \hat{\mathbf{C}}(\mathbf{r}_i) \right\rVert_2^2,
\end{equation}

where $\mathbf{C}(\mathbf{r}_i)$ is the color of pixel $i$ in the input images.

By replacing a traditional discrete volumetric representation with an MLP, NeRF makes a strong space-time trade-off: NeRF’s MLP requires multiple orders of magnitude less space than a dense voxel grid, but accessing the properties of the volumetric scene representation at any location requires an MLP evaluation instead of a simple memory lookup. Rendering a single ray that passes through the scene therefore requires querying the MLP many times.

\begin{figure*}[htbp]
    \centering
    \includegraphics[width=0.70\textwidth]{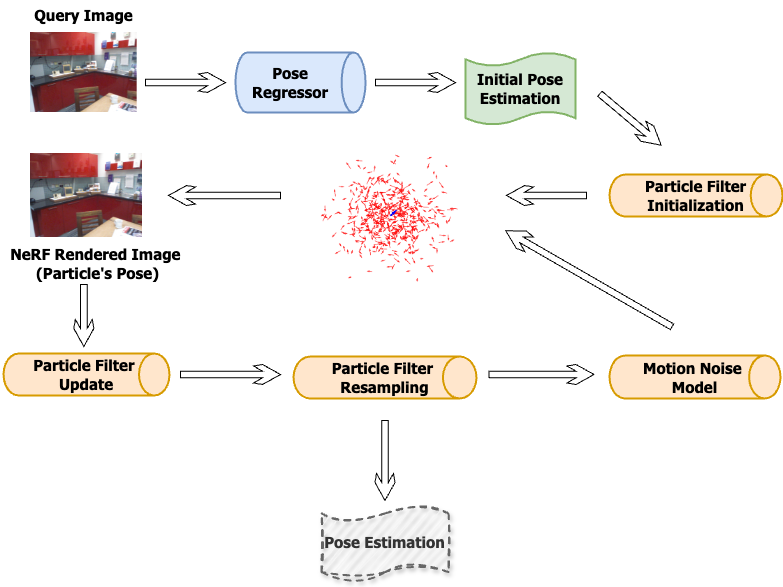}  
    \caption{HAL-NeRF Pipeline.}
    \label{fig:model}
\end{figure*}

\section{Method}
\label{method}

An overview of the proposed pipeline is presented in Fig \ref{fig:model}. It consists of two main components: pose regression and pose refinement. A Neural Radiance Field (NeRF) is trained to implicitly represent the scene. Using this trained NeRF, a set of synthetic images is generated alongside real observed images to train a pose regressor that predicts camera poses. In the first component, an initial pose estimate is obtained from a Convolution Neural Network regressor. In the second module, the pose is further refined using a Monte Carlo particle filter, inspired by the Loc-NeRF approach \cite{maggio2023loc}. The refinement process involves adjusting the particles' weights by comparing the photometric loss between the rendered images at each particle's pose and the query image, thereby enhancing the accuracy of the initial pose prediction.

\subsection{Pose Regressor}
\label{AA}

\subsubsection{NeRF Model}

Nerfacto \cite{tancik2023nerfstudio} is the NeRF implementation upon which our solutions concerning NeRFs are built. Nerfacto combines several ideas from NeRF research works and aims at a balance between rendering quality and computational demand. It is written in the nerfstudio python framework \cite{tancik2023nerfstudio} which is modular and aims to provide an easy-to-modify pipeline.

\subsubsection{Data Augmentation}

During the training phase of the pose regressor, additional training data can be generated by synthesizing multiple views from perturbed training poses. This process, referred to as Random View Synthesis (RVS), aims to enhance the network's ability to generalize to unseen perspectives.

Given an initial training pose $P$, a perturbed pose $P'$ is created by introducing random translation noise of $\psi$ meters and random rotation noise of $\phi$ degrees around the original pose. Subsequently, a synthetic image $I'$ is rendered using the Nerfacto model. The resulting synthetic pose-image pair ($P'$, $I'$) serves as an additional training sample for the pose estimator.

\subsubsection{Pose Regressor Architecture}

The described network is a deep learning architecture designed to perform both camera pose regression and feature extraction tasks, leveraging the well-known VGG-16 model \cite{simonyan2014very} as its backbone, based on DFNet \cite{chen2022dfnet}. The network integrates features from multiple layers to capture both low-level and high-level information, which is crucial for tasks like pose estimation. Specifically, the network extracts features from the $conv1_2$, $conv3_3$, and $conv5_3$ layers of VGG-16, and processes these through a series of adaptation layers that refine the extracted features using convolutional layers followed by ReLU activation and batch normalization. These features can be upsampled and returned as intermediate outputs.

For pose regression, a fully connected layer is employed after average pooling to predict the camera pose from the final feature map, making it suitable for tasks that require precise camera localization. The network's modular design allows it to output either the intermediate feature maps, the pose predictions, or both, depending on the task at hand. In this work, only the pose prediction module is used.

The loss function employed in this network is designed to ensure accurate camera pose predictions and robust feature learning by integrating three essential components: (a) the loss between the estimated pose derived from the ground truth image and the actual ground truth pose, denoted as $L_{gt}$, (b) the loss between the estimated pose derived from a rendered image, generated using the ground truth pose, and the actual ground truth pose, termed $L_{gt\_{rendered}}$, and (c) the loss between the estimated pose derived from a rendered image generated via Random View Synthesis (RVS) and a perturbed pose, labeled as $L_{pose\_{perturb}}$.

\begin{equation}
L = L_{gt} + L_{gt\_{rendered}} + L_{pose\_{perturb}}\label{eq}
\end{equation}

\subsection{Pose Optimization}
Once the initial pose prediction is obtained, particles are uniformly
initialized around this predicted pose. Subsequently, Monte Carlo NeRF localization is
utilized for high-accuracy pose estimation. By refining the initial pose prediction
from the regressor network with the iterative refinement capabilities of Loc-NeRF \cite{maggio2023loc}, the pipeline can achieve accurate localization results.

\begin{table*}[htbp]
\centering
\caption{Comparisons on 7-Scenes Dataset. \textnormal{We report the median translational and rotational errors in m/$^{\circ}$. The best results are highlighted in bold.}}
\label{tab:7scenes_comparison}
\begin{tabular}{|l|c|c|c|c|c|c|c|c|}
\hline
\textbf{Methods} & \textbf{Chess} & \textbf{Fire} & \textbf{Heads} & \textbf{Office} & \textbf{Pumpkin} & \textbf{Kitchen} & \textbf{Stairs} & \textbf{Average} \\
\hline
PoseNet & 0.10/4.02 & 0.27/10.0 & 0.18/13.0 & 0.17/5.97 & 0.19/4.67 & 0.22/9.51 & 0.35/10.5 & 0.21/7.74 \\
MapNet & 0.13/4.97 & 0.33/9.97 & 0.19/16.7 & 0.25/9.08 & 0.28/7.83 & 0.32/9.62 & 0.43/11.8 & 0.28/10.0 \\
MS-Transformer & 0.11/6.38 & 0.23/11.5 & 0.13/13.0 & 0.18/4.12 & 0.17/8.42 & 0.16/9.02 & 0.29/10.3 & 0.18/9.51 \\
PAE & 0.13/6.1 & 0.24/12.0 & 0.14/13.0 & 0.19/8.58 & 0.17/7.28 & 0.18/8.89 & 0.30/10.3 & 0.19/9.52 \\
DFNet & 0.03/1.12 & 0.06/2.30 & 0.04/2.29 & 0.06/1.54 & 0.07/1.92 & 0.07/1.74 & 0.12/2.63 & 0.06/1.93 \\
DFNet + NeFes$_{50}$ & 0.02/0.57 & 0.02/0.74 & 0.02/1.28 & 0.02/0.56 & 0.02/0.55 & 0.02/0.57 & 0.05/1.28 & 0.02/0.79 \\
\textbf{HAL-NeRF} & \textbf{0.003/0.33} & \textbf{0.008/0.6} & \textbf{0.013/0.92} & \textbf{0.008}/0.57 & 0.12/\textbf{0.35} & \textbf{0.005/0.5} & \textbf{0.015/0.87} & \textbf{0.025/0.59} \\
\hline
\end{tabular}
\end{table*}

\begin{table*}[h]
    \centering
    \caption{\textnormal{Percentage improvement in translational and rotational errors for each of the 7-Scenes after 50 iterations of the particle filter.}}
    \begin{tabular}{|c|c|c|}
        \hline
        \textbf{Scene} & \textbf{Translational Error Improvement (\%)} & \textbf{Rotational Error Improvement (\%)} \\
        \hline
        Chess       & 86.93\% & 72.35\% \\
        Fire        & 77.14\% & 67.72\% \\
        Heads       & 70.78\% & 68.07\% \\
        Office      & 71.98\% & 63.45\% \\
        Pumpkin     & 13.04\% & 68.73\% \\
        RedKitchen  & 85.91\% & 73.14\% \\
        Stairs      & 65.63\% & 39.09\% \\
        \hline
    \end{tabular}
    \label{tab:percentage_improvement}
\end{table*}

\subsubsection{Prediction Step}
    
In the prediction step of the particle filter, the set of particles $S_t$ at time $t$ is forecasted from the preceding set of particles $S_{t - 1}$ at time $t-1$ and from the motion model $O_t$.  In HAL-NeRF, since the task is pose refinement of a static image, a noise model was employed as the motion model. This noise model utilizes two zero-mean Gaussian distributions with relatively small standard deviations $\sigma_t$ (for translation) and $\sigma_r$ (for rotation). These small standard deviations were used to limit the exploration space with the assumption of a relatively accurate initial prediction, allowing for small adjustments as the refinement process progresses. This approach ensured that the refinement was focused on fine-tuning the pose rather than making large exploratory moves.

\subsubsection{Update Step}
    
In the update step, the query image $I_t$ is utilized to adjust the particles' weights $w^{i}_t$. Following standard Monte Carlo localization techniques, the weights are updated based on the measurement likelihood $P(I_t \mid X_i^t, M)$, which reflects the probability of capturing image $I_t$ from pose $X^{t}_i$ within the map M. To approximate this likelihood, a heuristic function \ref{eq:weight_update} is used, as in Loc-NeRF. This function evaluates the disparity between the query image $I_t$ and the image $C(r)$ projected by the NeRF map for each particle's pose $X^{t}_i$, assigning lower weights to particles that exhibit significant mismatches. To optimize computational efficiency, weights are updated using only a subset of M randomly sampled pixels from $I_t$ and $C(r)$.

\begin{equation}
    w_t^i = \left( \frac{M}{\sum_{j=1}^{M} (I_t(p_j) - C(r(p_j), X_i^t))^2} \right)^4
    \label{eq:weight_update}
\end{equation}

\subsubsection{Resampling Step}

Following the update step, resampling was performed by selecting $n$ particles from the
particle set $S_t$ with replacement. Each particle is sampled with a probability proportional to its weight $w^{i}_t$. This resampling process retains the particles that are more likely to represent accurate pose estimates while discarding less plausible hypotheses.

To enhance filter convergence and alleviate the computational burden, the number of particles $n$ and the standard deviations of the noise model's Gaussian distributions - $\sigma_t$ for translation and $\sigma_r$ for rotation - are dynamically adjusted as the process advances. This adjustment occurs when the standard deviation of the particles' translations falls below a specific threshold, indicating sufficient convergence. At this point, both the number of particles and the standard deviations are reduced, allowing the filter to focus more precisely on fine-tuning the pose. This approach not only reduces the search space but also optimizes computational resources, leading to a more efficient and accurate refinement process. Finally, the pose estimate $X_t$, is derived as a weighted average of the particle poses.

\section{Implementation Details}
\label{implementation}
\subsection{Implementation on 7-Scenes Dataset}
The 7-Scenes dataset \cite{shotton2013scene} is a widely used benchmark in the field of visual localization and pose estimation. It was created to evaluate the performance of camera relocalization algorithms, particularly in indoor environments. The dataset consists of RGB-D (color and depth) image sequences captured from seven different indoor scenes. These scenes are typically small, office-like environments with various objects, textures, and lighting conditions. 

As identified in \cite{brachmann2021limits}, the original ground truth RGB images and corresponding poses exhibit inconsistencies, leading to suboptimal NeRF renderings. To overcome this limitation, our method employed the COLMAP \cite{schonberger2016structure} structure-from-motion approach to extract camera poses, which provides more accurate results. The entire dataset was utilized for both training and testing, without applying any subsampling. All images were downsampled to a resolution of $160 \times 120$ pixels.

The pose regressor utilized a VGG-16 model pre-trained on ImageNet \cite{chen2022dfnet} as its backbone, and training was conducted using the Adam optimizer with a learning rate of $0.0001$. The Random View Synthesis module was applied every $25$ epochs, where all training samples were perturbed by a uniform random offset with a maximum radius of $0.1m$ for translation and $15^{\circ}$ for rotation. The integration of the Nerfacto model as the base NeRF model for this module has significantly improved the localization accuracy of the pose regressor. This enhancement is attributed to the higher quality of NeRF renderings, allowing the method to exploit a greater number of unseen views effectively.

\begin{figure*}[htbp]
    \centering
    \begin{subfigure}[b]{0.2\textwidth}
        \centering
        \includegraphics[width=\textwidth]{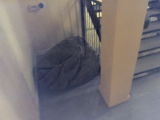}
        \caption{Query Image}
    \end{subfigure}
    \hspace{10pt}
    \begin{subfigure}[b]{0.2\textwidth}
        \centering
        \includegraphics[width=\textwidth]{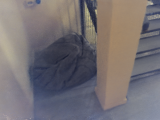}
        \caption{Step 1}
    \end{subfigure}
    \hspace{10pt}
    \begin{subfigure}[b]{0.2\textwidth}
        \centering
        \includegraphics[width=\textwidth]{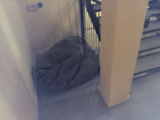}
        \caption{Step 50}
    \end{subfigure}
    
    \vspace{10pt}
    
    \begin{subfigure}[b]{0.2\textwidth}
        \centering
        \includegraphics[width=\textwidth]{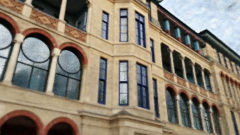}
        \caption{Query Image}
    \end{subfigure}
    \hspace{10pt}
    \begin{subfigure}[b]{0.2\textwidth}
        \centering
        \includegraphics[width=\textwidth]{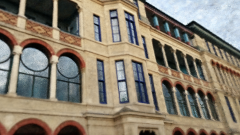}
        \caption{Step 1}
    \end{subfigure}
    \hspace{10pt}
    \begin{subfigure}[b]{0.2\textwidth}
        \centering
        \includegraphics[width=\textwidth]{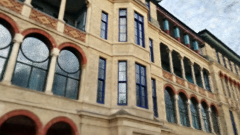}
        \caption{Step 50}
    \end{subfigure}

    \caption{Comparison of query images and refined results for Stairs (7-Scenes dataset) and Old Hospital (Cambridge dataset).}
    \label{fig:image_array}
\end{figure*}

\begin{figure*}[htbp]
    \centering
    \includegraphics[width=0.75\textwidth]{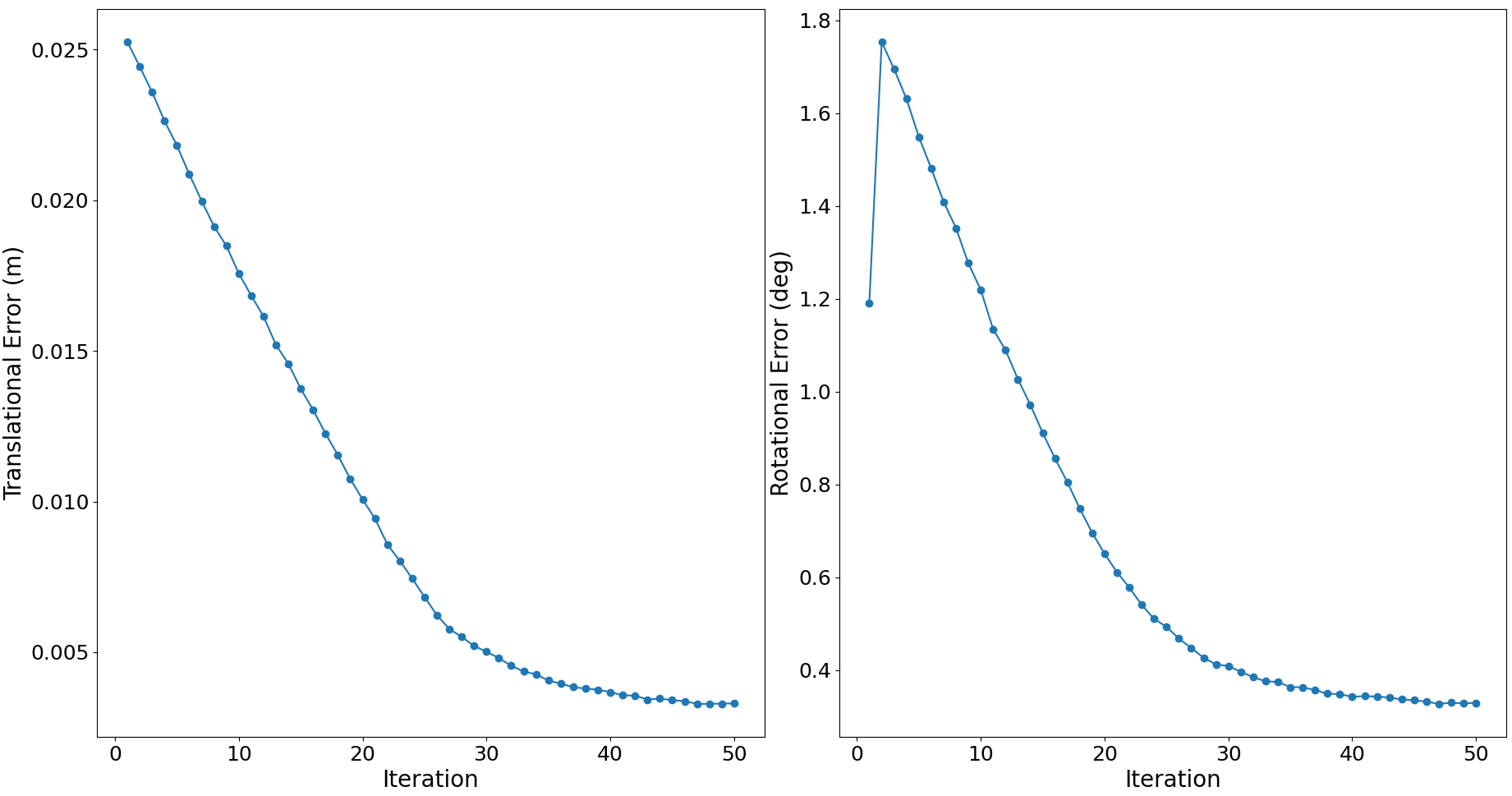}  
    \caption{Median translational and rotational errors for the Chess scene over 50 iterations.}
    \label{fig:Iterations}
\end{figure*}

For the pose refinement process, the Particle Filter employed $200$ particles in this experiment. All particles were initially distributed uniformly within a sphere of radius $0.02m$, centered around the initial pose estimation. This range was chosen to match the precision of the initial pose prediction. The motion noise model parameters were set to $0.005$ for both the translation and rotation components across all coordinates. When the standard deviation of the particles' translation falls below $0.01$, the alpha refinement mode is activated, reducing the number of particles to $100$. Additionally, the motion noise model parameters are halved. If the standard deviation further drops below $0.005$, the noise parameters are further reduced to one-fourth of their original values. The number of iterative steps for the pose refinement method was set to $50$. The choice of $50$ iterations was based on the observation that the median curves for both rotational and translational errors stabilize after $50$ iterations, as shown in Fig. \ref{fig:Iterations}. This indicates that further iterations would provide diminishing returns in terms of error reduction.

\subsection{Implementation on Cambridge Landmarks Dataset}

The Cambridge Landmarks dataset \cite{kendall2015posenet} is a comprehensive outdoor visual relocalization benchmark, captured in the region of Cambridge University. It includes handheld smartphone images of diverse scenes with substantial variations in lighting and exposure. The dataset covers expansive areas between $875m^2$ and $5600m^2$, offering a robust platform for evaluating visual localization methods.

COLMAP \cite{schonberger2016structure} software was employed to extract camera poses, which were then used as ground truth values. Since no models use COLMAP poses as ground truth on the Cambridge Landmarks dataset, we provide only a qualitative comparison with other models. The entire training and testing datasets, across various sequences, were utilized in our experiments. Additionally, all images were downsampled to a resolution of $240 \times 135$ to standardize the input data. 

The training parameters for the pose regressor were kept consistent with those used for the $7$ Scenes dataset, with the exception of the Random View Synthesis module. In this case, the rotation perturbation component was set to $5^\circ$.

The parameters for the particle filter were kept identical to those used for the 7-Scenes dataset, with the exception of the rotational motion noise, which was adjusted to $0.01$.

\begin{figure}[htbp]
    \centering
    \begin{subfigure}[b]{0.35\textwidth}
        \centering
        \includegraphics[width=\textwidth]{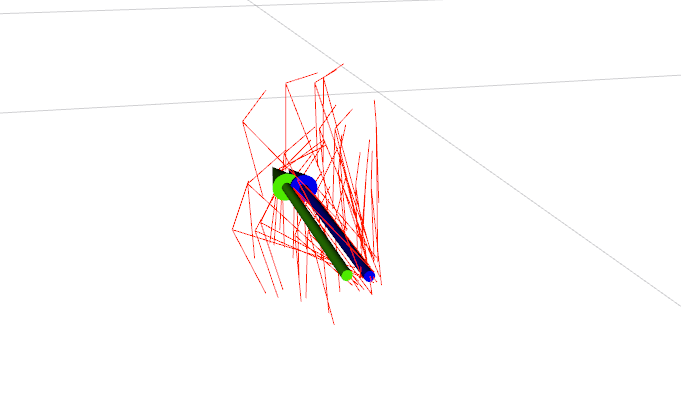} 
        \caption{Particle Initialization}
        \label{fig:step1}
    \end{subfigure}
    \hspace{15pt} 
    \begin{subfigure}[b]{0.35\textwidth}
        \centering
        \includegraphics[width=\textwidth]{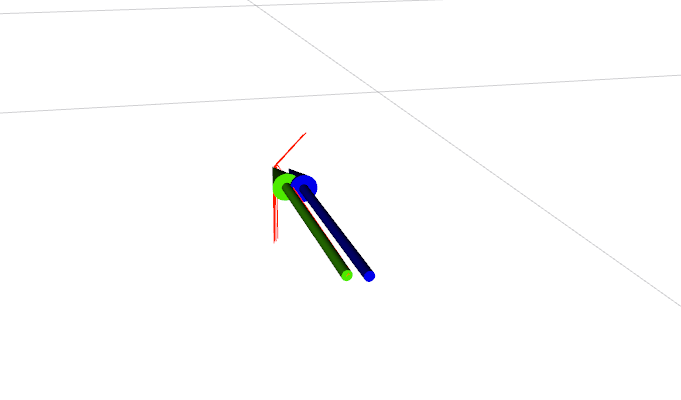} 
        \caption{Convergence at $50^{th}$ Iteration}
        \label{fig:step50}
    \end{subfigure}
    
    \caption{Visualization of particle convergence in the \texttt{rviz} environment from the $1^{st}$ to the $50^{th}$ iteration step.}
    \label{fig:particle_convergence}
\end{figure}

\section{Results}
\label{results}
\subsection{Evaluation on 7-Scenes Dataset}

\begin{table*}[htbp]
\centering
\caption{Comparisons on Cambridge Landmarks. \textnormal{Median translational and rotational errors in m/$^{\circ}$. The best results are highlighted in bold.}}
\label{tab:cambridge_landmarks_comparison}
\begin{tabular}{|l|c|c|c|c|c|}
\hline
\textbf{Methods} & \textbf{Kings} & \textbf{Hospital} & \textbf{Shop} & \textbf{Church} & \textbf{Average} \\
\hline
PoseNet (PN) & 1.66/4.86 & 2.62/4.90 & 1.41/7.18 & 2.45/7.96 & 2.04/6.23 \\
MapNet & 1.07/1.89 & 1.94/3.91 & 1.49/4.22 & 2.00/4.53 & 1.63/3.64 \\
MS-Transformer & 0.83/1.47 & 1.81/2.39 & 0.86/3.07 & 1.62/3.99 & 1.28/2.73 \\
DFNet & 0.73/2.37 & 2.00/2.98 & 0.67/2.21 & 1.37/4.03 & 1.19/2.90 \\
DFNet + NeFeS$_{50}$ & 0.37/0.54 & 0.52/0.88 & 0.15/0.53 & 0.37/1.14 & 0.35/0.77 \\
\textbf{HAL-NeRF (COLMAP)} & \textbf{0.06}/0.70 & \textbf{0.09/0.53} & \textbf{0.01/0.26} & \textbf{0.03/0.85} & \textbf{0.04/0.58} \\
\hline
\end{tabular}
\end{table*}

As illustrated in Table~\ref{tab:7scenes_comparison}, our model demonstrates superior performance compared to the referenced models that also utilized the same COLMAP poses. The only experiment where translation estimates underperformed was on the Pumpkin scene. This was due to the lower quality of the extracted NeRF model for this specific case, which resulted in suboptimal data augmentation. Consequently, the initial pose estimation and refinement process was less effective compared to other scenes. Table \ref{tab:percentage_improvement} presents the percentage improvement in both translational and rotational errors from the $1^{st}$ iteration to the $50^{th}$ iteration for each scene. This demonstrates the effectiveness of the particle filter refinement process across the 7-Scenes dataset. 

In the first row of Fig. \ref{fig:image_array}, images rendered at the $1^{st}$ and $50^{th}$ iteration steps alongside the query image for the Stairs scene, are presented. In the 7-Scenes dataset, where the initial pose estimation is already highly accurate, the visual perception between the $1^{st}$ and $50^{th}$ iteration steps may appear subtle. However, a detailed view demonstrates the method's capability for achieving high-precision localization and fine-tuning both more and less accurate initial estimations.

\subsection{Evaluation on Cambridge Landmarks Dataset}

The comparison of our model with other approaches is presented in Table \ref{tab:cambridge_landmarks_comparison}. As the other models did not use COLMAP poses as ground truth, the comparison is indicative. However, the results demonstrate strong performance, highlighting the effectiveness of our approach. 

Fig.~\ref{fig:particle_convergence} illustrates the convergence of particles from the $1^{st}$ to the $50^{th}$ iteration for an image from the Old Hospital dataset. Initially, the particles (red) are distributed around the pose prediction (blue) and progressively refine their positions towards the ground truth pose (green).

\section{Discussion}
\label{discussion}

The primary objective of this paper was to leverage Nerfacto's advanced rendering capabilities for data augmentation during pose regressor training and for employing a Particle Filter as a refinement technique to improve the initial pose estimation, ultimately achieving high-precision localization. 

The first module of the pipeline achieved a fine initial pose estimation, which is also attributed to the data augmentation with Nerfacto. Possibly, rendered views from another model could provide even better training examples.

The second refinement module significantly reduced errors in all cases, though it came at the expense of computational speed.  As Particle Filters inherently distribute particles within an exploration space, the accurate initial prediction allowed us to limit the exploration space around the estimated pose. Since the query image was likely to be near this estimation, a broader exploration space was not required. Despite this,  leveraging the limited size of the exploration space by reducing the number of particles and thus processing time, was beneficial up to a limit. The number of particles was initially set to $200$ and gradually reduced to $100$ through the particle annealing module, but further reduction proved ineffective, leading to a decline in performance. The processing time of the refinement module was proportional to the number of particles. $200$ particles demanded almost twice the time of $100$ particles. Therefore, modifications that do not require a large number of particles are essential. The other factors that consume time are the number of iterations required to converge ($50$) and the rendering time of the NeRF model. In future work, modifications to reduce iterations, using momentum, larger update step and/or particle filtering specific operations will be addressed. Besides, NeRF rendering will be accelerated. 

Another aspect of the method is its relation to a NeRF model. For the particle filter to function effectively, high-quality renderings must be generated to ensure that the update function operates meaningfully. While Nerfacto performs well in most cases, the method is prone to failure if a scene cannot be accurately modeled as a high-quality NeRF. Therefore, high-quality NeRF is essential but also a parallel task. The digital twins applications demand photorealistic and fast rendering. For dynamic XR environments, occlusions of transient objects and appearance differences must be taken into account.

Furthermore, the limited exploration space results in similar RGB renderings from the poses of the distributed particles. To improve the efficiency of the weight update process, the update function could incorporate not only the photometric difference but also an additional parameter that captures depth or semantic information from the scene, which is already embedded in the weights of the NeRF.

\section{Conclusion}
\label{conclusion}

In this paper, we presented HAL-NeRF, a novel high-accuracy camera localization method that leverages the capabilities of Neural Radiance Fields (NeRFs) combined with a Monte Carlo particle filter for iterative pose refinement. By utilizing the advanced Nerfacto model for data augmentation and fine-tuning, we achieved significant improvements in localization accuracy over traditional Absolute Pose Regression (APR) methods. Our experiments on the 7-Scenes and Cambridge Landmarks datasets demonstrated state-of-the-art performance, highlighting the effectiveness of integrating NeRF-based representations into the localization pipeline. Despite the computational demands of the particle filter, the proposed approach successfully addresses the challenges of accurate pose estimation in complex scenes. Future work will focus on optimizing the computational efficiency and extending the method to handle more dynamic and larger-scale environments.

\end{document}